\documentclass[letterpaper]{article}
\usepackage{aaai25}
\usepackage{amssymb}
\usepackage{booktabs}
\usepackage{times}
\usepackage{helvet}
\usepackage{courier}
\usepackage{multirow}
\usepackage[hyphens]{url}
\usepackage{graphicx}
\usepackage{natbib}
\usepackage{caption}
\usepackage{tikz}
\usetikzlibrary{arrows.meta,positioning,shapes.geometric}

\usepackage{amsthm}
\theoremstyle{definition}
\newtheorem{exmp}{Example}

\title{Language Models Do Not Embed Numbers Continuously}
\author {
    Alex O. Davies\textsuperscript{\rm 1,\rm 2},
    Roussel Nzoyem\textsuperscript{\rm 1,\rm 2},
    Nirav Ajmeri\textsuperscript{\rm 2},
    Telmo M. Silva Filho\textsuperscript{\rm 2}
}
\affiliations {
    \textsuperscript{\rm 1}Primary Author\\
    \textsuperscript{\rm 2}University of Bristol, UK\\
    alexander.davies@bristol.ac.uk, rd.nzoyemngueguin@bristol.ac.uk
}

\begin{document}

\maketitle

\begin{abstract}
Recent research has extensively studied how large language models manipulate integers in specific arithmetic tasks, and on a more fundamental level, how they represent numeric values.
These previous works have found that language model embeddings can be used to reconstruct the original values, however, they do not evaluate whether language models actually model continuous values \textit{as} continuous.
Using expected properties of the embedding space, including linear reconstruction and principal component analysis, we show that language models not only represent numeric spaces as non-continuous but also introduce significant noise.
Using models from three major providers (OpenAI, Google Gemini and Voyage AI), we show that while reconstruction is possible with high fidelity ($R^2 \geq 0.95$), principal components only explain a minor share of variation within the embedding space.
This indicates that many components within the embedding space are orthogonal to the simple numeric input space.
Further, both linear reconstruction and explained variance suffer with increasing decimal precision, despite the ordinal nature of the input space being fundamentally unchanged.
The findings of this work therefore have implications for the many areas where embedding models are used, in-particular where high numerical precision, large magnitudes or mixed-sign values are common.

\end{abstract}

\section{Introduction}

Large Language Models (LLMs), trained on next token prediction over internet-wide data, demonstrate extraordinary emergent abilities to manipulate numbers and perform arithmetic operations beyond their training date.
For this reason, they are increasingly deployed in complex safety-critical scenarios requiring complex mathematical reasoning, such as accounting \cite{yoo2024much}, medical calculations \cite{khandekar2024medcalc}, radiotherapy planning \cite{wang2025feasibility}, to name but a few.

These deployments pose serious safety concerns, emphasising the need to investigate how LLMs represent numbers. 
One problem which plagues both the expressivity and computational efficiency of LLMs is the long-range dependency \cite{vaswani2017attention,gu2023mamba}.
It is commonly understood that LLMs perform better which shorter prompts, or focus on a specific parts of the prompt when it is too long \cite{hengle2024multilingual}.
That said, when inserting decimal numbers in LLM prompts, users tend to include arbitrary number of decimal places, thereby prolonging the size of length of the prompt.
Knowing how well LLMs represent numbers based on their precision stands to empower both users and practitioners.

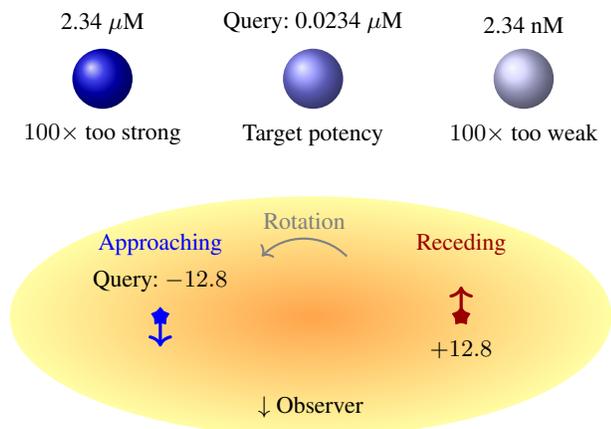
\begin{figure}[t]
\centering
\begin{tikzpicture}[scale=0.8]
  
  
  \shade[ball color=blue!90!black] (1.5,2) circle (0.5);
  \node[above, font=\small] at (1.5,2.6) {2.34 $\mu$M};
  \node[below, font=\footnotesize] at (1.5,1.4) {$100\times$ too strong};
  
  \shade[ball color=blue!50] (5,2) circle (0.5);
  \node[above, font=\small] at (5,2.6) {Query: 0.0234 $\mu$M};
  \node[below, font=\footnotesize] at (5,1.4) {Target potency};
  
  \shade[ball color=blue!20] (8.5,2) circle (0.5);
  \node[above, font=\small] at (8.5,2.6) {2.34 nM};
  \node[below, font=\footnotesize] at (8.5,1.4) {$100\times$ too weak};
  
  
\end{tikzpicture}

\vspace{1.5em}

\begin{tikzpicture}[scale=0.8]
  
  \shade[inner color=orange, outer color=yellow!50, opacity=0.7] (5,3) ellipse (5 and 2);
  
  \draw[->, thick, gray] (5.6,4) arc (45:135:1) node[midway, above, font=\small] {Rotation};

  \node[blue, font=\small] at (2.5,4.2) {Approaching};
    \draw[->, blue, very thick] (2.5,3) -- (2.5,2.5);
  
  \node[red!60!black, font=\small] at (7.5,4.2) {Receding};
  \draw[->, red!60!black, very thick] (7.5,3) -- (7.5,3.5);
  

  \node[star, star points=5, fill=blue, draw=blue, minimum size=0.24cm, inner sep=0pt] at (2.5,3) {};
  \node[above, font=\small] at (2.5,3.25) {Query: $-12.8$};

  \node[star, star points=5, fill=red!60!black, draw=red!60!black, minimum size=0.24cm, inner sep=0pt] at (7.5,3) {};
  \node[below, font=\small] at (7.5,2.75) {$+12.8$};
  
  
  \node[font=\small] at (5,1.5) {$\downarrow$ Observer};
\end{tikzpicture}

\caption{Visualisations of our examples for LLM embeddings in scientific knowledge applications. \textbf{Top:} In material science, the magnitude of a concentration is crucial, but repeated mantissas in numerical representations could cause incorrect retrieval. \textbf{Bottom:} In astronomy, a negatively signed value may not indicate it is semantically \textit{opposite} to its positive counterpart, such as in measuring velocities within galactic disks.}
\label{fig:scientific-examples}
\end{figure}

The same models are also applied in scientific domains.
In some works embedding models are used as-is; \citet{Peikos2024} and \citet{Amugongo2025} use multiple models for medical document retrieval, both applications that require a complex understanding of numerical values.
Other works fine-tune over domain-specific data before using a model's embeddings, such as \citet{Lin2024} for geoscience applications and \citet{Choudhary2024} for material property prediction and retrieval.
Again, these works assume that LLMs are capable of usefully understanding continuous numerical values, despite significant differences in properties or semantic meaning across a range of numerical values.
We provide examples from different scientific fields, with visualisations in Figure~\ref{fig:scientific-examples}:

\begin{exmp}[Climate Science]
    Atmospheric aerosol concentrations can range from $10^{-12}$kg m$^{-3}$ in exceptionally clean air (e.g. the Arctic) to $10^{-3}$kg m$^{-3}$ in urban areas. A scientist querying for \textit{``Black carbon concentrations around 2.847 × 10$^{-9}$kg m$^{-3}$''} might receive ``2.847 × 10$^{-6}$ kg m$^{-3}$'' ranked as highly similar simply because both strings share the mantissa 2.847.
\end{exmp}

\begin{exmp}[Drug Discovery]
    Inhibitor potencies span many orders of magnitude: picomolar compounds (10$^{-12}$ M) are ultra-tight binders suitable for therapeutic development, while millimolar compounds (10${-3}$ M) bind so weakly they're considered inactive.
    A medicinal chemist searching for \textit{``IC50 values near 0.0234 $\mu$M''} might retrieve compounds at 2.34 nM or 2.34 $\mu$M as similar matches.
\end{exmp}    

There are also examples of where users would expect a more complex -- and not strictly linear -- interpretation of continuous values: 

\begin{exmp}[Astronomy]
    Stellar velocities within a galactic disk might range from -500 kms$^{-1}$ (stars moving toward the observer) to +500 km$^{-1}$ (stars moving away), with precision to 0.001 km$^{-1}$ required to detect exoplanets via Doppler wobble.
    An astronomer querying for "velocity -12.847 km$^{-1}$" might not retrieve "+12.847 km$^{-1}$" as similar, despite the physical equivalency on opposite sides of the galactic disk, as the magnitude is opposite.
\end{exmp}

The behaviours these models exhibit in embedding numerical values is therefore crucial before application.
A model which embeds strictly linearly is useful in the climate and material science examples, but would require more complex treatment where magnitude does not necessarily imply dis-similarity.







Since other recent works have highlighted representational space as critical for arithmetic \cite{maltoni2024arithmetic,zhu2024language}, a plethora of methods have attempted to understand the semantics of the LLM embedding space.
The current literature is heavily focused on the representation and processing of integers within specific arithmetic or reasoning tasks \cite{levy2024language,kantamneni2025language}.
Some studies have attempted to reframe numerical representations to improve LLM performance \cite{schwartz2024numerologic,zhang2024reverse}, while others have brought attention to the connections to human cognition \cite{shah2023numeric,alquboj2025number}.
While this has yielded fascinating insights into the complex, often non-linear geometry of numerical representations, it leaves a more fundamental question unanswered: \emph{how well do embedding models encode the basic semantic value of continuous real numbers across varying scales and precisions?}

Our work addresses this gap by providing a \textbf{general} and \textbf{lightweight} framework for evaluating the semantic fidelity of numerical embeddings.
Rather than decoding a specific geometric structure (e.g., a helix or circle \cite{kantamneni2025language}), we propose a set of metrics (namely linear $R^2$, PCA correlation and explained variance) that directly quantify how well an embedding captures the one-dimensional nature of a scalar value.
Using these metrics we provide a critical insight: the complex, multi-faceted representations identified in other studies (string-entanglements, periodic features) manifest as quantifiable ``noise'' in the embedding space.
Our methodology thus offers a scalable and task-agnostic tool to measure the purity and robustness of any model's ability to represent the simple concept of numerical magnitude.

Specifically, our contributions are as follows:
\begin{enumerate}
    \item[(1)] A general and lightweight framework to evaluate the fidelity of continuous embeddings, addressing a critical gap in the literature in a model-agnostic manner. 
    
    \item[(2)] Task-independent metrics (linear $R^2$, PCA correlation, and explained variance) which demonstratively quantify how well an embedding captures the ordinal, one-dimensional nature of the scalar values. These quantify the fidelity of numerical representations, enabling practitioners to better understand model limitations and optimize prompt design.
    
    \item[(3)] A breadth of experiment to validate our framework and the proposed metrics. We go beyond the scope of the current literature by considering positive decimals, mixed sign decimals, and mixed sign integers.
\end{enumerate}

\section{Related Work}


A central challenge for Large Language Models (LLMs) is their inconsistent and often fragile ability to perform numerical reasoning.
Recent research has tackled this from multiple angles: improving arithmetic performance through novel processing techniques, probing the internal geometry of numerical representations, and drawing parallels between model and human numerical cognition.

\paragraph{Internal Representations and Processing Methods}
The current theme in the literature is that LLMs introduce complex patterns into embeddings for simple 1D scalar values \cite{zhu2024language}.
Probing experiments reveal highly complex internal structures.
For instance, numbers appear to be encoded using per-digit circular representations in base 10 \cite{levy2024language}, which helps explain why model errors are often digit-based rather than value-based.
For specific operations like addition, models have been shown to develop even more intricate structures, representing numbers as a generalized helix and manipulating them with a trigonometric ``Clock'' algorithm \cite{kantamneni2025language}.
While limited to whole number representations, \citet{kantamneni2025language} show that LLMs contain the ability to abstract away the continuous space.
\citet{zhou2024pre} identify Fourier features with varying levels of periodicity within these representations, which are later used for arithmetic operations.
This internal complexity is further compounded by a fundamental ambiguity: LLM representations are often an entanglement of a number's value and its string-like properties, where similarity is influenced by various metrics of distance \cite{marjieh2025number}.

\paragraph{Reformatting Numerical Representations:} To improve performance in light of these complex representations, researchers have focused on reformatting inputs to align with computational logic.
NumeroLogic \cite{schwartz2024numerologic}, for example, prefixes numbers with their digit count to provide essential place-value context upfront.
In a similar vein, Little-Endian Fine-Tuning (LEFT) \cite{zhang2024reverse} reverses the order of digits to mimic human-like computation (least significant digit first), dramatically improving efficiency and accuracy in arithmetic tasks.
Neither work, unlike ours, investigates the role of number magnitude and precision.
The need for these techniques is underscored by comprehensive benchmarks which reveal that modern LLMs still fail at a wide range of basic numerical tasks.
For example, the NUPA Test \cite{yang2024number} shows broad failures beyond simple addition, while \citet{tang2025investigating} highlight significant error rates in the seemingly straightforward task of numerical translation, especially when dealing with large units across languages.



\section{Experiments}


Consider a real scalar number $x \in \mathbb{R}^1$, and an embedding model $f(x) \rightarrow \hat{x} \in \mathbb{R}^d$ for $d$ the dimensionality of the model's embedding space.
Scalars $x$ are in a set of $X = \{x_1, x_2,\ldots\}$.
Further, consider that $x \in X$ has a given number of integer and decimal places $a$ and $b$;

\begin{table}[h]
\centering
\begin{tabular}{c p{3pt} c}
    1234 & . & 567 \\
    \_ \_ \_ \_ & . & \_  \_ \_ \\
    $a=4$ & & $b=3$\\
\end{tabular}
\end{table}

\begin{figure}[ht]
\centering
\begin{tikzpicture}[x=0.08\linewidth, y=0.05\linewidth]
    \node[draw, rounded corners, minimum width=1.5cm, minimum height=0.7cm] (input) at (0,0) {\small $X \in \mathbb{R}^1$};
    \node[above] at (0,1.2) {\footnotesize scalars};
    
    \node[draw, rounded corners, minimum width=1.5cm, minimum height=0.7cm] (embedding) at (3.5,0) {\small $\hat{X} \in \mathbb{R}^d$};
    \node[above] at (3.5,1.2) {\footnotesize embeddings};
    
    \node[draw, rounded corners, minimum width=1.5cm, minimum height=0.7cm] (linear) at (7,1) {\footnotesize lin$(\hat{X})$};
    \node[draw, rounded corners, minimum width=1.5cm, minimum height=0.7cm] (pca) at (7,-1) {\footnotesize PCA$(\hat{X})$};
    
    \draw[->, thick] (input) to node[above] {\footnotesize $f(x)$} (embedding);
    
    \draw[->, thick] (embedding.east) to[out=30, in=180] (linear.west);
    \draw[->, thick] (embedding.east) to[out=-30, in=180] (pca.west);

    \node (lin-corr) at (10,1) {\footnotesize Eqn.~\ref{eqn:linear-corr}};

    \node (pca-corr) at (10,-0.5) {\footnotesize Eqn.~\ref{eqn:expl-pca}};

    \node (pca-var) at (10,-1.5) {\footnotesize Eqn.~\ref{eqn:corr-pca}};

    \draw[->, thick] (linear.east) to[out=0, in=180] (lin-corr.west);

    \draw[->, thick] (pca.east) to[out=0, in=180] (pca-corr.west);

    \draw[->, thick] (pca.east) to[out=0, in=180] (pca-var.west);
    
\end{tikzpicture}
    \caption{Framework for measuring numerical embedding quality. Scalars are embedded into high-dimensional space and evaluated using linear reconstruction and PCA to quantify preservation of numerical structure through three complementary metrics, defined in Equations~(\ref{eqn:linear-corr},\ref{eqn:expl-pca},\ref{eqn:corr-pca}).}
    \label{fig:setup}
\end{figure}
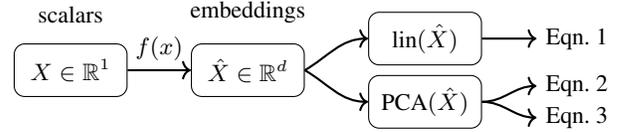

In this work we evaluate how accurately embedding models encode numbers $x$ with respect to these precisions $a,b$ and sign of the number.
A diagram of our experimental framework is presented in Figure~\ref{fig:setup}.
We can expect that, as in prior work \cite{zhu2024language, levy2024language}, a number $x$ can be reproduced by a linear model over the embedding $\textrm{lin}(\hat{X}) \rightarrow X'$; perfect reconstruction is $\textrm{lin}(\hat{X}) = X$.
More precisely, we expect that there is very good correlation between predicted values from this linear model $X'$ and the original scalars $X$.

\begin{equation}
    \textrm{corr}(X', X) \simeq 1
    \label{eqn:linear-corr}
\end{equation}

Further, we know that there is only one component of variation, with $\textrm{Rank}(X) = 1$.
As a result we expect that strong embeddings of numbers, with `understanding' of numeric spaces, would similarly have only one component of variation, $\textrm{Rank}(\hat{X}) \simeq 1$.
As a result, Principal Component Analysis (PCA) over the embedded scalars $\hat{X}$ should have an explained variance ratio in the first component of approximately 1:

\begin{equation}
    \textrm{VR} = \frac{\lambda_0}{\sum_{0}^{d} \lambda_i} \simeq 1
    \label{eqn:expl-pca}
\end{equation}

with $\lambda_i$ the eigenvalues of the covariance matrix.
We can also expect that the primary direction of variation in the embedding space is in-line with the original set of scalars $X$:

\begin{equation}
    \textrm{corr}(\textrm{PCA}_0, X) \simeq 1.
    \label{eqn:corr-pca}
\end{equation}

For a perfect encoder, and continuous internal representations, we expect the relationships in Equations~\ref{eqn:linear-corr},\ref{eqn:expl-pca} and \ref{eqn:corr-pca} to hold true regardless of integer and decimal places $a,b$ or the signs of the numbers.

\begin{figure*}[hbtp]
    \centering
    \includegraphics[width=0.9\linewidth]{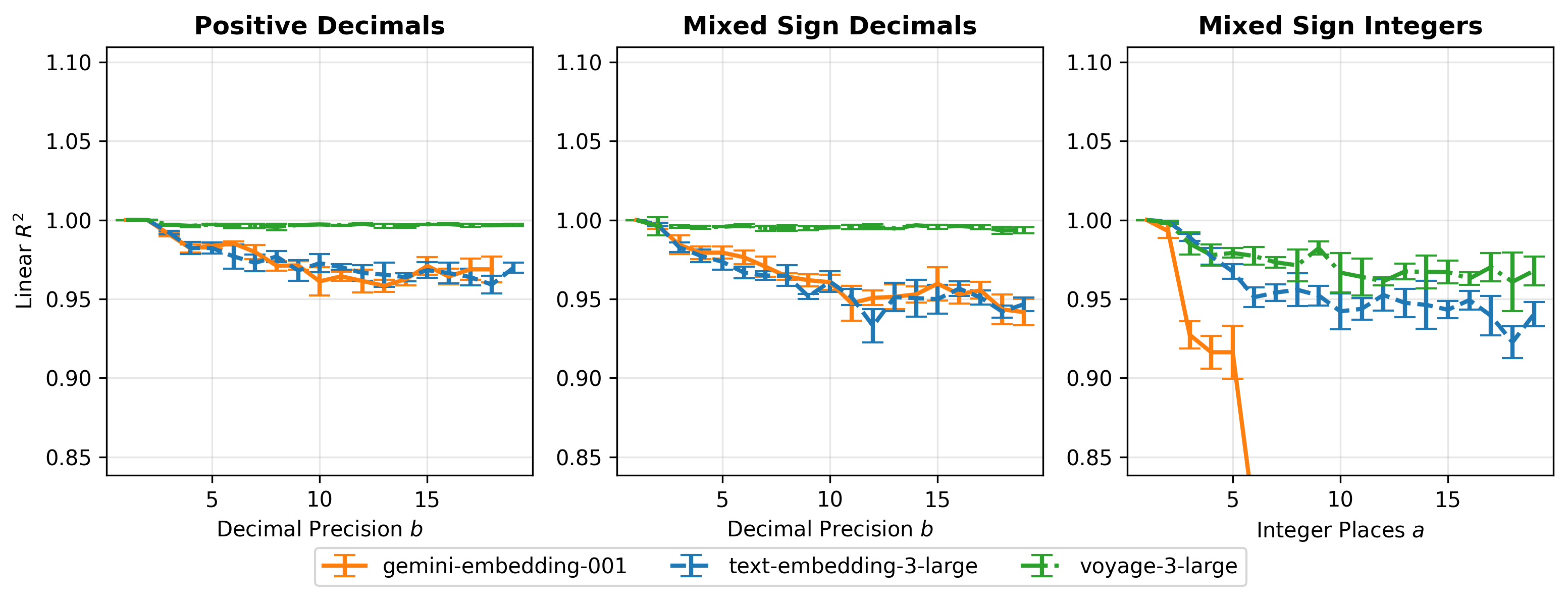}
    \caption{Decimal precision for each dataset plotted against the $R^2$ score of the linear model reconstructing the original scalars $X$ from their embedded counterparts $\hat{X}$.}
    \label{fig:lin-corr}

    \centering
    \includegraphics[width=0.9\linewidth]{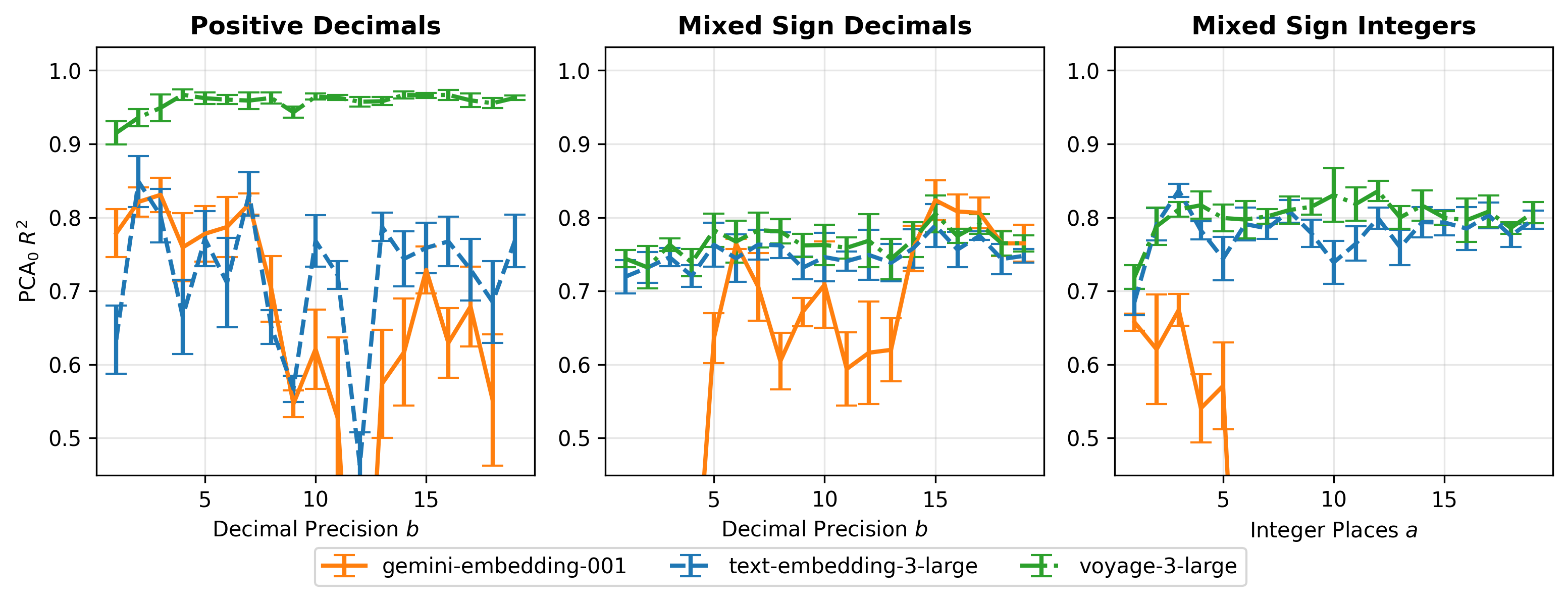}
    \caption{Decimal precision for each dataset plotted against the $R^2$ of the first component of a PCA projection of the embedded samples $\hat{X}$ against their original counterparts $\hat{X}$.}
    \label{fig:pca-corr}

    \centering
    \includegraphics[width=0.9\linewidth]{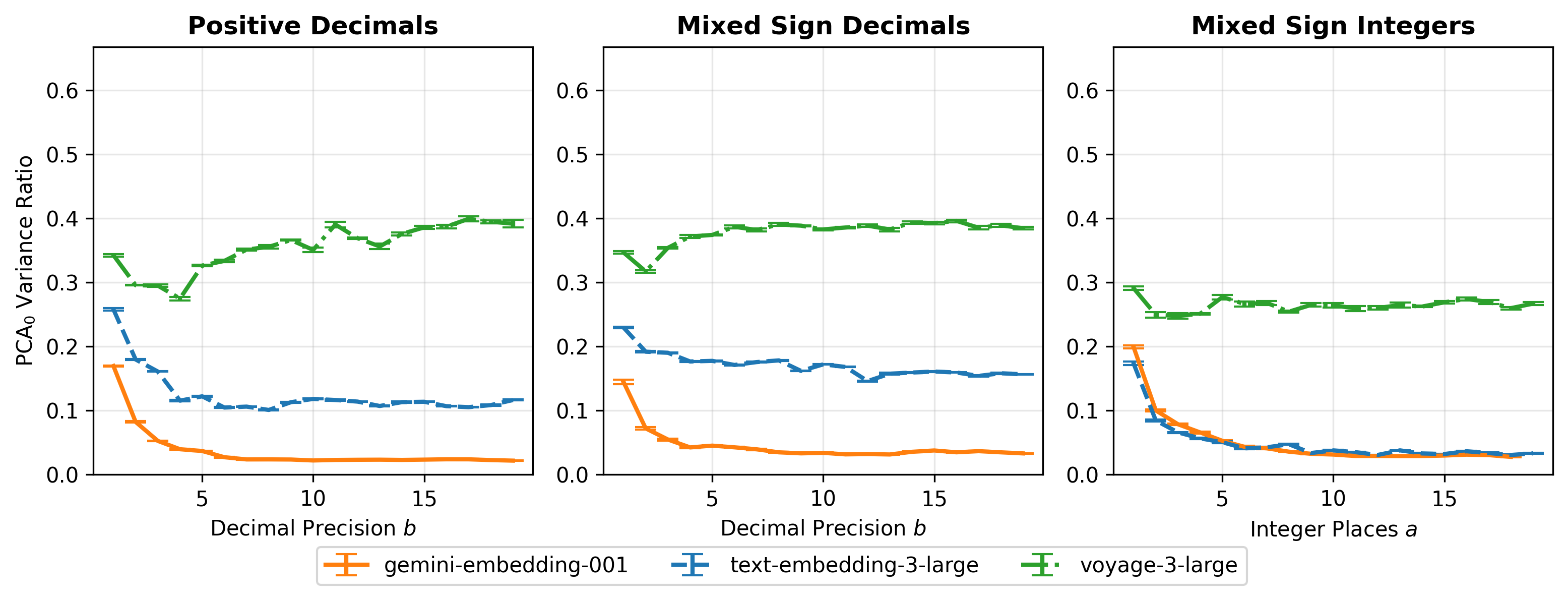}
    \caption{Decimal precision for each dataset plotted against the explained variance ratio of the first component of a PCA projection of the embedded samples $\hat{X}$.}
    \label{fig:pca-var}
\end{figure*}

\subsubsection{Datasets}

We produce three datasets of scalars $X$.
The first is positive decimals $x \in [0,1]$ with $a=1$ and with the precision of the decimals iteratively increased from $a = 1$ to $a=20$.
The second is mixed sign decimals $x \in [-1,1]$ with the same precisions as the positive decimals.
The third is mixed-sign scalars of varied integer and zero decimal places $b=0$, $a \in [0,20]$.
The range of this dataset varies according to $a$, $-10^{(a)} < x < 10^{(a)}$.
Across the three datasets individual scalars are sampled randomly, up to 500 samples per dataset.
We adopt a 5-fold split to produce error margins.

\subsubsection{Embedding Models}

In our evaluation of LLMs as embedding models for numeric data, we use models which are specifically targetted at embedding applications.
We use all of the embedding models available from three public providers:

\begin{description}
    \item[Gemini]\footnote{\url{https://ai.google.dev/gemini-api/docs/embeddings}}Provided by Google, we use the \texttt{gemini-embedding-001} model with the default (largest) embedding dimension.
    Gemini embedding models are initialised from the Gemini LLM, with two rounds of contrastive training to produce the embedding model \cite{lee2025}.
    Gemini embedding models are reported to out-perform competitors on benchmarks.

    \item[OpenAI]\footnote{\url{https://platform.openai.com/docs/guides/embeddings}} we use the three latest models provided by OpenAI, namely \texttt{text-embedding-3-large}, a condensed version \texttt{text-embedding-3-large} and the older\texttt{text-embedding-ada-002}.
    No companion paper is published, but public documentation and releases suggests that OpenAI's embedding models are based on the GPT-4 family, and are contrastively fine-tuned for embeddings \cite{OpenAI}.
    OpenAI models are reported to out-perform competitors on benchmarks.
    

    \item[VoyageAI]\footnote{\url{https://docs.voyageai.com/docs/embeddings}}, provided by MongoDB, provides several text embedding models. We evaluate their non-specialist models, namely the 3.5 series (default, \texttt{lite} and \texttt{large}).
    No companion paper is published, with public documentation suggesting pure contrastive training for these embedding models \cite{Voyage}.
    VoyageAI models are reported to out-perform competitors on benchmarks.
\end{description}

Quantitative results for each dataset and metric can be found in Table~\ref{tab:positive_decimals} for positive decimals, in Table~\ref{tab:mixed_sign_decimals} for mixed-sign decimals, and in Table~\ref{tab:mixed_int_decimal} for high-magnitude mixed-sign integers.

\begin{table*}[h]
\caption{Metrics for linear models and principal components over a dataset of positive numbers of varying decimal places.}
\label{tab:positive_decimals}
\small
\centering
\begin{tabular}{llccccccc}
\toprule
Model & Provider & \multicolumn{2}{c}{Linear $R^2$} & \multicolumn{2}{c}{PCA $R^2$} & \multicolumn{2}{c}{PCA Variance} \\
\cline{3-8}
 &  & Min & Max & Min & Max & Min & Max \\
\midrule
gemini-embedding-001 & Google & 0.96 $\pm$ 0.00 & 1.00 $\pm$ 0.00 & 0.04 $\pm$ 0.04 & 0.83 $\pm$ 0.02 & 0.03 $\pm$ 0.00 & 0.20 $\pm$ 0.00 \\
text-embedding-3-large & OpenAI & 0.96 $\pm$ 0.01 & 1.00 $\pm$ 0.00 & 0.46 $\pm$ 0.04 & 0.85 $\pm$ 0.03 & 0.03 $\pm$ 0.00 & 0.17 $\pm$ 0.00 \\
text-embedding-3-small & OpenAI & 0.96 $\pm$ 0.01 & 1.00 $\pm$ 0.00 & 0.64 $\pm$ 0.04 & 0.87 $\pm$ 0.03 & 0.04 $\pm$ 0.00 & 0.19 $\pm$ 0.00 \\
text-embedding-ada-002 & OpenAI & 0.95 $\pm$ 0.01 & 1.00 $\pm$ 0.00 & 0.25 $\pm$ 0.03 & 0.85 $\pm$ 0.02 & 0.04 $\pm$ 0.00 & 0.17 $\pm$ 0.00 \\
voyage-3-large & Voyage & 1.00 $\pm$ 0.00 & 1.00 $\pm$ 0.00 & 0.92 $\pm$ 0.02 & 0.97 $\pm$ 0.01 & 0.25 $\pm$ 0.00 & 0.29 $\pm$ 0.00 \\
voyage-3.5 & Voyage & 0.97 $\pm$ 0.01 & 1.00 $\pm$ 0.00 & 0.02 $\pm$ 0.02 & 0.93 $\pm$ 0.01 & 0.14 $\pm$ 0.00 & 0.36 $\pm$ 0.00 \\
voyage-3.5-lite & Voyage & 0.97 $\pm$ 0.01 & 1.00 $\pm$ 0.00 & 0.64 $\pm$ 0.02 & 0.86 $\pm$ 0.01 & 0.18 $\pm$ 0.00 & 0.33 $\pm$ 0.00 \\
\bottomrule
\end{tabular}
\end{table*}

\begin{table*}[h]
\caption{Metrics for linear models and principal components over a dataset of numbers of varying decimal places and mixed signs.}
\label{tab:mixed_sign_decimals}
\small
\centering
\begin{tabular}{llccccccc}
\toprule
Model & Provider & \multicolumn{2}{c}{Linear $R^2$} & \multicolumn{2}{c}{PCA $R^2$} & \multicolumn{2}{c}{PCA Variance} \\
\cline{3-8}
 &  & Min & Max & Min & Max & Min & Max \\
\midrule
gemini-embedding-001 & Google & 0.94 $\pm$ 0.01 & 1.00 $\pm$ 0.00 & 0.01 $\pm$ 0.01 & 0.82 $\pm$ 0.03 & 0.03 $\pm$ 0.00 & 0.14 $\pm$ 0.00 \\
text-embedding-3-large & OpenAI & 0.93 $\pm$ 0.01 & 1.00 $\pm$ 0.00 & 0.72 $\pm$ 0.02 & 0.79 $\pm$ 0.03 & 0.15 $\pm$ 0.00 & 0.23 $\pm$ 0.00 \\
text-embedding-3-small & OpenAI & 0.87 $\pm$ 0.01 & 1.00 $\pm$ 0.00 & 0.71 $\pm$ 0.02 & 0.78 $\pm$ 0.03 & 0.18 $\pm$ 0.00 & 0.30 $\pm$ 0.00 \\
text-embedding-ada-002 & OpenAI & 0.85 $\pm$ 0.03 & 1.00 $\pm$ 0.00 & 0.69 $\pm$ 0.04 & 0.77 $\pm$ 0.02 & 0.13 $\pm$ 0.00 & 0.21 $\pm$ 0.00 \\
voyage-3-large & Voyage & 0.99 $\pm$ 0.00 & 1.00 $\pm$ 0.00 & 0.73 $\pm$ 0.03 & 0.80 $\pm$ 0.02 & 0.32 $\pm$ 0.00 & 0.40 $\pm$ 0.00 \\
voyage-3.5 & Voyage & 0.96 $\pm$ 0.01 & 1.00 $\pm$ 0.00 & 0.73 $\pm$ 0.02 & 0.79 $\pm$ 0.02 & 0.38 $\pm$ 0.00 & 0.45 $\pm$ 0.00 \\
voyage-3.5-lite & Voyage & 0.91 $\pm$ 0.11 & 1.00 $\pm$ 0.00 & 0.75 $\pm$ 0.02 & 0.80 $\pm$ 0.02 & 0.37 $\pm$ 0.00 & 0.55 $\pm$ 0.00 \\
\bottomrule
\end{tabular}
\end{table*}

\begin{table*}[h]
\caption{Metrics for linear models and principal components over a dataset of numbers of varying integer places with mixed signs.}
\label{tab:mixed_int_decimal}
\small
\centering
\begin{tabular}{llccccccc}
\toprule
Model & Provider & \multicolumn{2}{c}{Linear $R^2$} & \multicolumn{2}{c}{PCA $R^2$} & \multicolumn{2}{c}{PCA Variance} \\
\cline{3-8}
 &  & Min & Max & Min & Max & Min & Max \\
\midrule
gemini-embedding-001 & Google & 0.48 $\pm$ 0.08 & 0.99 $\pm$ 0.01 & -0.03 $\pm$ 0.02 & 0.72 $\pm$ 0.02 & 0.02 $\pm$ 0.00 & 0.10 $\pm$ 0.00 \\
text-embedding-3-large & OpenAI & 0.91 $\pm$ 0.02 & 1.00 $\pm$ 0.00 & 0.69 $\pm$ 0.03 & 0.84 $\pm$ 0.02 & 0.10 $\pm$ 0.00 & 0.19 $\pm$ 0.00 \\
text-embedding-3-small & OpenAI & 0.89 $\pm$ 0.01 & 1.00 $\pm$ 0.00 & 0.41 $\pm$ 0.07 & 0.82 $\pm$ 0.01 & 0.07 $\pm$ 0.00 & 0.20 $\pm$ 0.00 \\
text-embedding-ada-002 & OpenAI & 0.90 $\pm$ 0.01 & 1.00 $\pm$ 0.00 & 0.59 $\pm$ 0.03 & 0.79 $\pm$ 0.01 & 0.09 $\pm$ 0.00 & 0.17 $\pm$ 0.00 \\
voyage-3-large & Voyage & 0.95 $\pm$ 0.02 & 1.00 $\pm$ 0.01 & 0.73 $\pm$ 0.01 & 0.83 $\pm$ 0.02 & 0.28 $\pm$ 0.00 & 0.44 $\pm$ 0.00 \\
voyage-3.5 & Voyage & 0.93 $\pm$ 0.01 & 1.00 $\pm$ 0.00 & 0.67 $\pm$ 0.02 & 0.85 $\pm$ 0.02 & 0.29 $\pm$ 0.00 & 0.38 $\pm$ 0.00 \\
voyage-3.5-lite & Voyage & 0.90 $\pm$ 0.06 & 1.00 $\pm$ 0.00 & 0.68 $\pm$ 0.02 & 0.87 $\pm$ 0.02 & 0.26 $\pm$ 0.00 & 0.42 $\pm$ 0.00 \\
\bottomrule
\end{tabular}
\end{table*}


\subsection{Linear Reconstruction}

Here we evaluate the relationship in Equation~\ref{eqn:linear-corr}, that is, a linear model can perfectly reconstruct samples $X$ from the corresponding samples in the embedding space $\hat{X}$.
Figure~\ref{fig:lin-corr} shows linear $R^2$ scores plotted against size $a$ and $b$ for each dataset.

On positive decimals performance for all models degrades as the precision ($b$) increases, though most models maintain $\textrm{corr}(X, X') \geq 0.95$, indicating that the original samples can be well-constructed from the embedding space even at very high precisions.
Introducing mixed-sign decimals degrades performance for all models, most notably for condensed models.
OpenAI models in-particular degrade in performance far more with decimal precision when negative decimals are included.
Allowing integer places, and larger magnitudes, leads to greater degradation from all models.
In-particular, Google's Gemini-based model drops to at-best medium correlation with precision beyond $a=b=7$.

Overall we echo the findings of prior work that LLM embeddings of numbers can broadly be used to reconstruct those numbers.
However, when numbers are allowed to range in sign and magnitude, such reconstruction suffers.
Notably, for true `understanding' of the simple numeric space our models are encoding, there should be little to no variation in these correlations with integer or decimal places.

\subsection{PCA Correlation}


Next, to evaluate the preservation of Equation~\ref{eqn:corr-pca}, we measure the correlation of the first PCA component PCA$_0$ over the embedded datasets $\hat{X}$ against the original scalars they represent.
This measures, in effect, whether the embedding correctly encodes the direction of the input samples.
Perfect encoders would lead to perfect correlation between PCA$_0$ and $X$.
We visualise these results in Figure~\ref{fig:pca-corr}.

Performance on positive decimals is highly volatile, with most models dipping with increasing precision into low correlations.
Only \texttt{voyage-3-large} maintains high correlations against increasing decimal precision, though all models at one decimal place are at least fairly well correlated with the original samples.
On both the larger magnitude and mixed sign decimals the Gemini model again performs poorly, though on the mixed sign dataset it recovers some performance with increasing precision.
Notably all other models perform comparably on the mixed sign decimal dataset, with little overall degradation as precision increases, and at roughly the same correlation values as on the positive decimals dataset.
We discuss these results in more detail in the Discussion, but overall we have shown that the principle component of the embedded samples contains at least most of the information necessary to reconstruct the ordering of $X$.

\subsection{PCA Explained Variance}

\begin{figure}[ht]
    \centering
    \includegraphics[width=\linewidth]{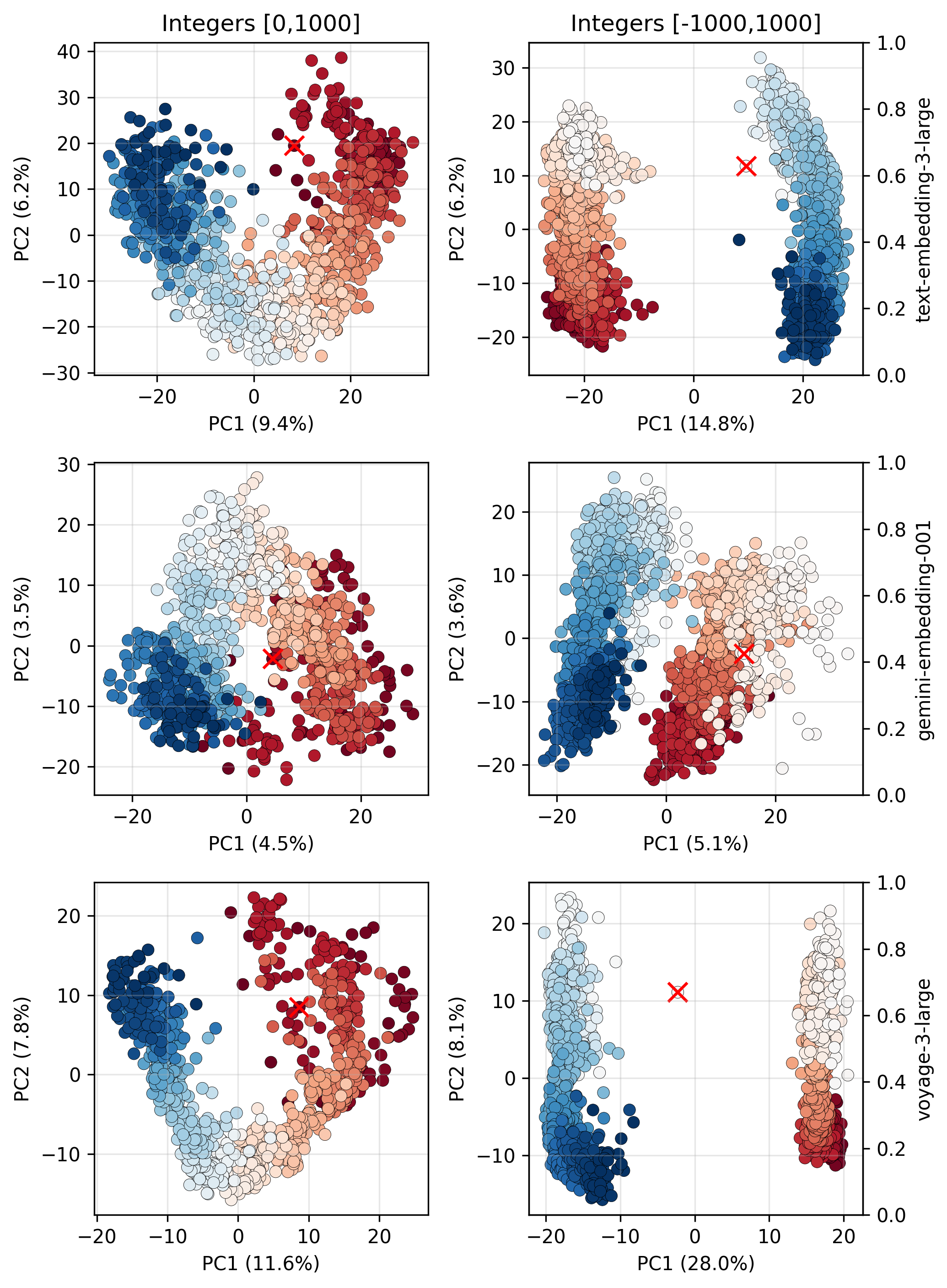}
    \caption{Visualisations of the first two principal components of embeddings of the integers $x \in [0,1000]$ (left) and $x \in [-1000,1000]$ (right) for the main OpenAI, Voyage and Gemini models. `$\times$' symbols mark $x=0$.}
    \label{fig:pca}
\end{figure}

Finally we evaluate the preservation of Equation~\ref{eqn:expl-pca}, measuring the relative amount of variance explained by the first principle component PCA$_0$.
We expect, given the one-dimensional and uniform nature of the input samples $X$, that PCA$_0$ explains \textit{all} of the variance in the embedded samples $\hat{X}$.
We visualise the explained variance ratio for each model and dataset against increasing precision in Figure~\ref{fig:pca-var}.

On positive decimals all models show the same exponential-like explained variance decrease with increasing decimal precision, with at most 40\% of variance explained at one decimal place.
Results are more spread on the mixed sign dataset, with explained variances increasing for most models, with the same pattern on the larger magnitude integer and decimal dataset.
Voyage models overall explain more variance in their first principle component, and actually increase in this explained variance as precision increases.
Models from OpenAI and Google simply decrease in performance with increasing precision, with the Gemini model explaining the least variance in its first principle component.

\section{Discussion}




Our results reveal a complex picture of how LLM embedding models encode numerical information.
First, we demonstrate that embeddings can indeed be used to reconstruct numbers with reasonable fidelity.
The linear reconstruction experiments show that most models maintain $R^2$ scores above 0.95 for simpler numerical datasets, confirming that numerical information is preserved in the embedding space.
This finding supports prior work suggesting that language models possess some inherent understanding of numerical relationships.

Second, the first principal component of the embedded representations correlates meaningfully with input precision across most models and datasets.
This correlation indicates that the primary axis of variation in the embedding space aligns with the numerical ordering of the input scalars, suggesting that models do capture the fundamental ordinal structure of numbers.

However, our third finding reveals a significant limitation: the explained variance by the first principal component remains consistently low across all models, typically below 40\% even for the simplest datasets.
This low explained variance is particularly concerning when considered alongside our fourth observation.
Given that the first component does correlate well with the original numbers, the low explained variance implies that the embedding space contains substantial additional variation that is not present in the original one-dimensional numerical input.

In Figure~\ref{fig:pca} we visualise the first two principal components of these principal components for the `flagship' model from each provider over the complete sets of integers $x_{+} \in [0,1000]$ and $x_{\pm} \in [-1000,1000]$.
$x_+$ shows that while the first principal component does broadly encode the rank of the original data, the second principal component introduces clear trends unrelated to the original values.
Further, all three models on $x_{\pm}$ have the first principal component encoding effectively only the sign of the data, despite the fundamentally continuous nature of the original values.
The second component then broadly encodes magnitude, although for the gemini model there is significant `bleed' between the clusters for positive and negative values.
In the Appendix we reproduce Figure~\ref{fig:pca} for increasing number magnitude, and observe that for large magnitudes neither the first nor second principal components represent number magnitude.

This excess variation suggests that embedding models introduce considerable noise into their numerical representations.
The high-dimensional embedding spaces capture not only the intended numerical information but also artifacts from the models' pretraining on diverse text corpora.
These artifacts manifest as spurious dimensions of variation that obscure the underlying numerical structure.

\subsection{Implications}

Our findings have several practical implications for applications utilizing LLM embeddings for numerical data.
First, numerical understanding appears to degrade significantly with increasing precision, suggesting that simply rounding numbers to fewer decimal places may improve performance in downstream tasks.
This finding is particularly relevant for applications requiring numerical reasoning or similarity computation over quantitative data.

Second, the signs of numbers have substantial impact on embedding quality across all tested models.
The principal component of variation for all models, with mixed sign values, comes to represent only the sign of the numbers (see Figure~\ref{fig:pca}).
The introduction of negative values consistently degrades performance metrics, indicating that models struggle with the concept of negative numbers more than might be expected.
This limitation suggests caution when using embeddings for datasets containing both positive and negative values.

Third, embedding models introduce systematic noise into numerical representations, with those based on large language models being particularly prone to this issue.
The low explained variance ratios demonstrate that much of the embedding space is devoted to capturing information orthogonal to the numerical content itself.
This noise may interfere with applications requiring precise numerical relationships, such as mathematical reasoning or quantitative analysis tasks.

\section{Conclusion}

We have conducted a comprehensive evaluation of numerical precision in LLM embedding models, examining how well these systems encode scalar values across different ranges and precisions.
Our analysis reveals that while embedding models can preserve numerical information sufficiently for linear reconstruction, they introduce substantial noise that limits their effectiveness for precise numerical applications.

The key finding is that embedding models exhibit a fundamental trade-off between preserving numerical information and introducing extraneous variation.
While the primary component of variation in embeddings does correlate with numerical values, the majority of the embedding space encodes information unrelated to the numerical content.
This suggests that current embedding models, despite their success in many natural language processing tasks, may not be optimal for applications requiring precise numerical understanding.

Future work should focus on developing embedding architectures specifically designed for numerical data, potentially through specialized pretraining objectives or architectural modifications that better isolate numerical information from other sources of variation.
Additionally, investigating techniques for denoising numerical embeddings or identifying the most relevant dimensions for numerical tasks could improve the practical utility of existing models for quantitative applications.








\clearpage
\newpage

\appendix

\begin{figure*}[hbtp]
    \centering
    \includegraphics[width=0.8\textwidth]{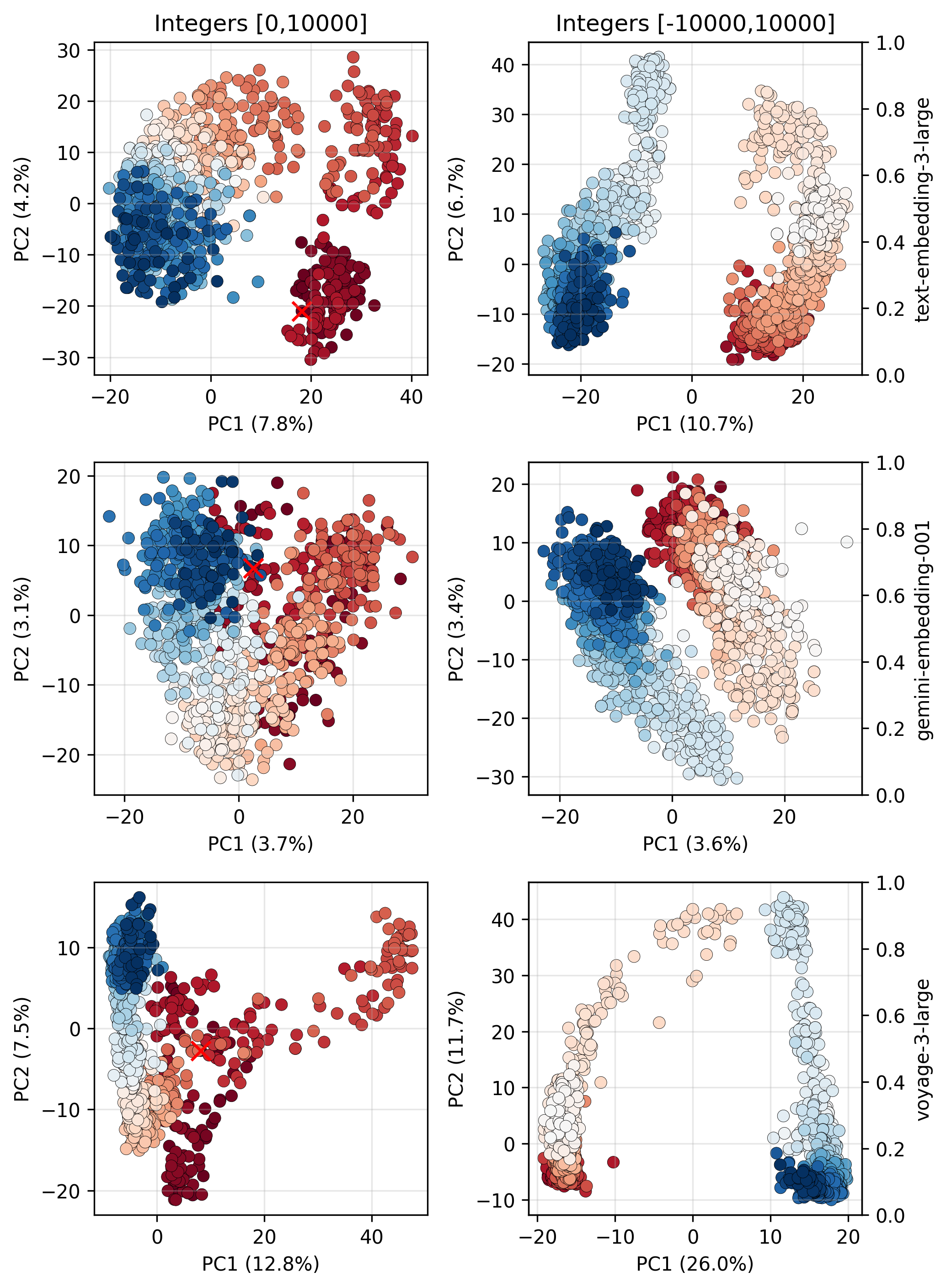}
    \caption{PCA over randomly sampled $x \in [0,10\textrm{k}], |X|=1000$ (left) and $x \in [-10\textrm{k},10\textrm{k}], |X|=2000$ (right) for the main OpenAI, Voyage and Gemini models.}
    \label{fig:pca-10k}
\end{figure*}

\begin{figure*}[hbtp]
    \centering
    \includegraphics[width=0.8\textwidth]{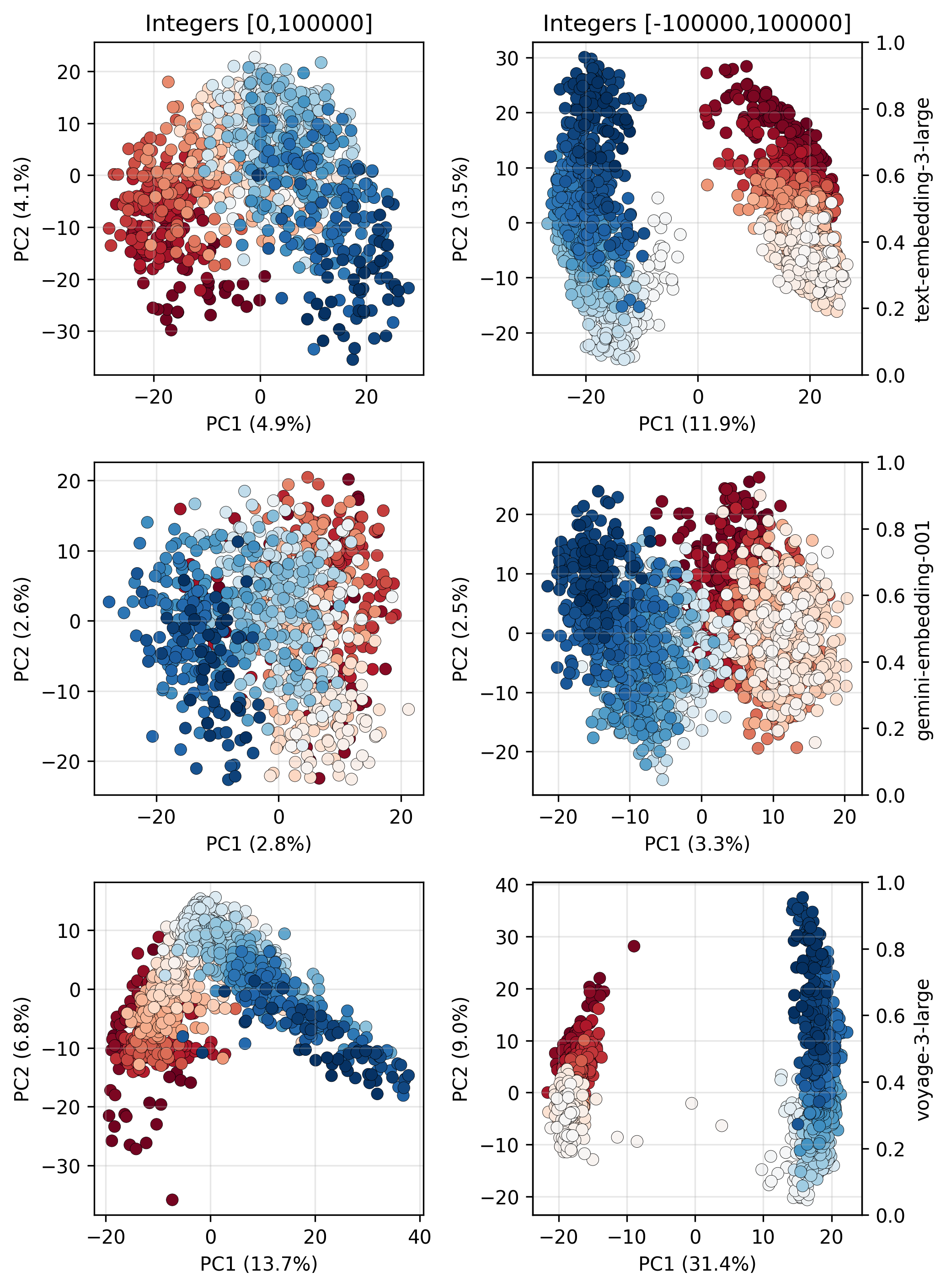}
    \caption{PCA over randomly sampled $x \in [0,100\textrm{k}], |X|=1000$ (left) and $x \in [-100\textrm{k},100\textrm{k}], |X|=2000$ (right) for the main OpenAI, Voyage and Gemini models.}
    \label{fig:pca-1M}
\end{figure*}

\begin{figure*}[hbtp]
    \centering
    \includegraphics[width=0.8\textwidth]{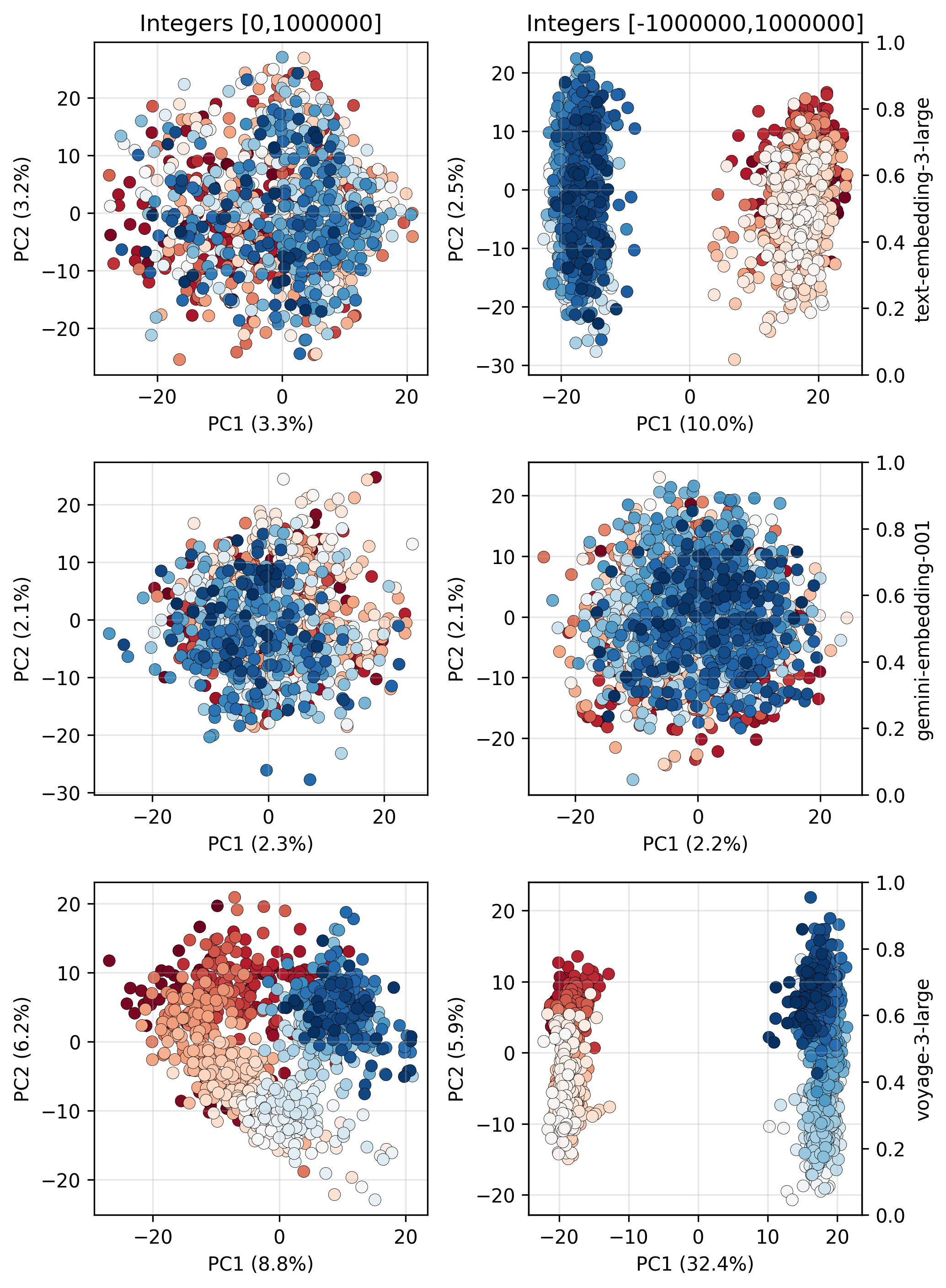}
    \caption{PCA over randomly sampled $x \in [0,1\textrm{M}], |X|=1000$ (left) and $x \in [-1\textrm{M},1\textrm{M}], |X|=2000$ (right) for the main OpenAI, Voyage and Gemini models.}
    \label{fig:pca-100k}
\end{figure*}

\begin{figure*}[hbtp]
    \centering
    \includegraphics[width=0.8\textwidth]{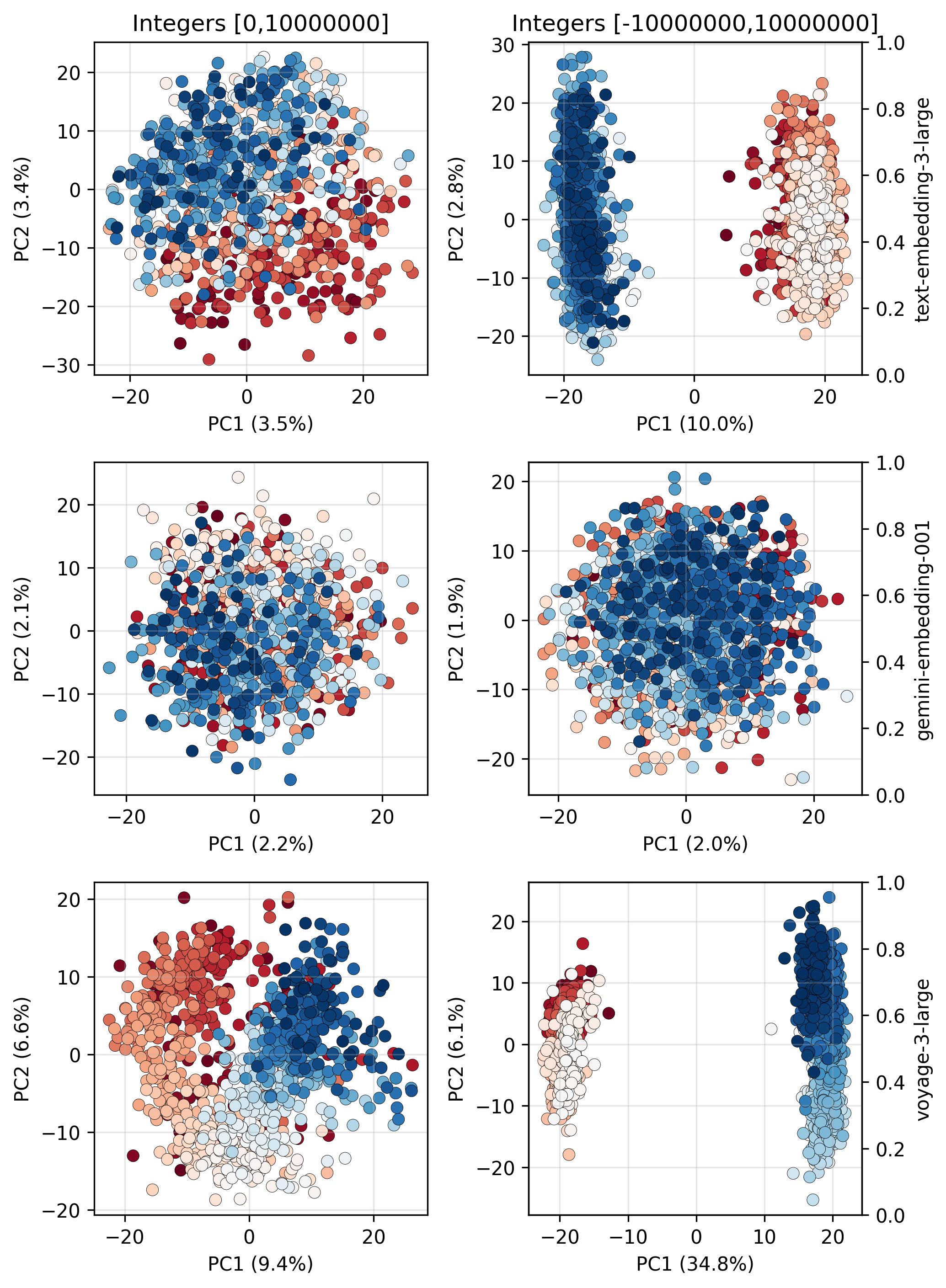}
    \caption{PCA over randomly sampled $x \in [0,10\textrm{M}], |X|=1000$ (left) and $x \in [-10\textrm{M},10\textrm{M}], |X|=2000$ (right) for the main OpenAI, Voyage and Gemini models.}
    \label{fig:pca-10M}
\end{figure*}

\end{document}